\documentclass{article}

\usepackage[preprint]{corl_2026} 

\usepackage{booktabs} 
\usepackage{graphicx}
\usepackage{amsmath}
\usepackage{xspace}
\usepackage{booktabs} 
\usepackage{multirow}
\usepackage{tcolorbox}
\tcbuselibrary{breakable, skins}
\usepackage{enumitem}
\usepackage{listings}
\usepackage{wrapfig}
\usepackage{makecell}
\usepackage{longtable}
\usepackage{subcaption}
\usepackage{capt-of}

\newtcolorbox{promptbox}[1][]{
  colback=blue!5,
  colframe=blue!40,
  fonttitle=\bfseries\ttfamily\small,
  fontupper=\small,
  breakable,
  #1
}

\title{A Few Words Go a Long Way: \\
Language Guided Robot Policy Synthesis}

\author{
  \mdseries Daphne Chen$^{1}$\thanks{Correspondence to \texttt{daphc@cs.washington.edu}} \quad Archit Ritesh Jain$^{1}$ \quad
  Eric Goossen$^{1}$ \quad Emma Romig$^{1}$ \\
  Michael Murray$^{2}$ \quad Nick Walker$^{3}$ \quad Maya Cakmak$^{1}$ \\[4pt]
  $^{1}$University of Washington \quad
  $^{2}$Microsoft Research \quad
  $^{3}$Massachusetts Institute of Technology \\[4pt]
}

\newcommand{\method}{\textsc{ARCHITECT}\xspace}

\begin{document}
\maketitle


\begin{abstract}
    While vision-language-action models have demonstrated impressive zero-shot manipulation capabilities, they remain fundamentally black box policies that are difficult to interpret, adapt, or correct when they inevitably fail. In this work, we propose \method, a framework that treats robot policy acquisition as an interactive program synthesis task. \method leverages the reasoning capabilities of LLM coding agents to synthesize modular robot programs that utilize a suite of perception and control tools. Unlike end-to-end models where distribution shift leads to unpredictable, cascading failures, our modular architecture allows users to isolate failures and localize feedback at the level of abstraction required. We introduce an iterative process where a human supervisor provides natural language corrections to steer the policy. These corrections are grounded in the policy code by program execution traces and distilled into a persistent skill library, a form of long-term in-context learning which enables the agent to accumulate a repertoire of reusable, interpretable behaviors. In a benchmark evaluation on a Franka Panda robot, \method outperforms state-of-the-art VLA models and program synthesis baselines on complex, long-horizon tasks, including articulated object manipulation and cloth folding. Our results demonstrate that the synthesized skill library enables the system to transfer to novel tasks with decreasing human intervention, providing a steerable and data-efficient alternative to black-box robot learning. Website: \url{https://robo-architect.github.io/}
\end{abstract}



\section{Introduction}

    Vision-Language-Action (VLA) models have demonstrated remarkable breadth across robotic manipulation tasks, but they suffer from a significant adaptability gap: they are inherently uninterpretable and require massive datasets to rectify even minor behavioral errors. In real-world deployment, slight distribution shifts, such as a new object or a small perturbance in the scene, can cause VLAs to fail. Small failures cascade as the robot moves further out of distribution. Furthermore, these black-box architectures lack the structural priors necessary for complex, multi-stage reasoning and are difficult to adapt to novel constraints or failure modes without expensive fine-tuning. Even with parameter-efficient methods, adapting a 7B-parameter VLA requires hours of GPU time on dedicated hardware for as few as a thousand demonstrations, and full foundation model training runs cost orders of magnitude more \citep{kim2024openvlaopensourcevisionlanguageactionmodel}. For a human user, there is no way to ``steer'' a black-box VLA policy other than collecting more expert demonstrations and retraining, a process which is prohibitively expensive for most real-world users \citep{maslej2025artificialintelligenceindexreport}. 

    \begin{figure}[!t]
        \centering
        \includegraphics[width=0.8\textwidth]{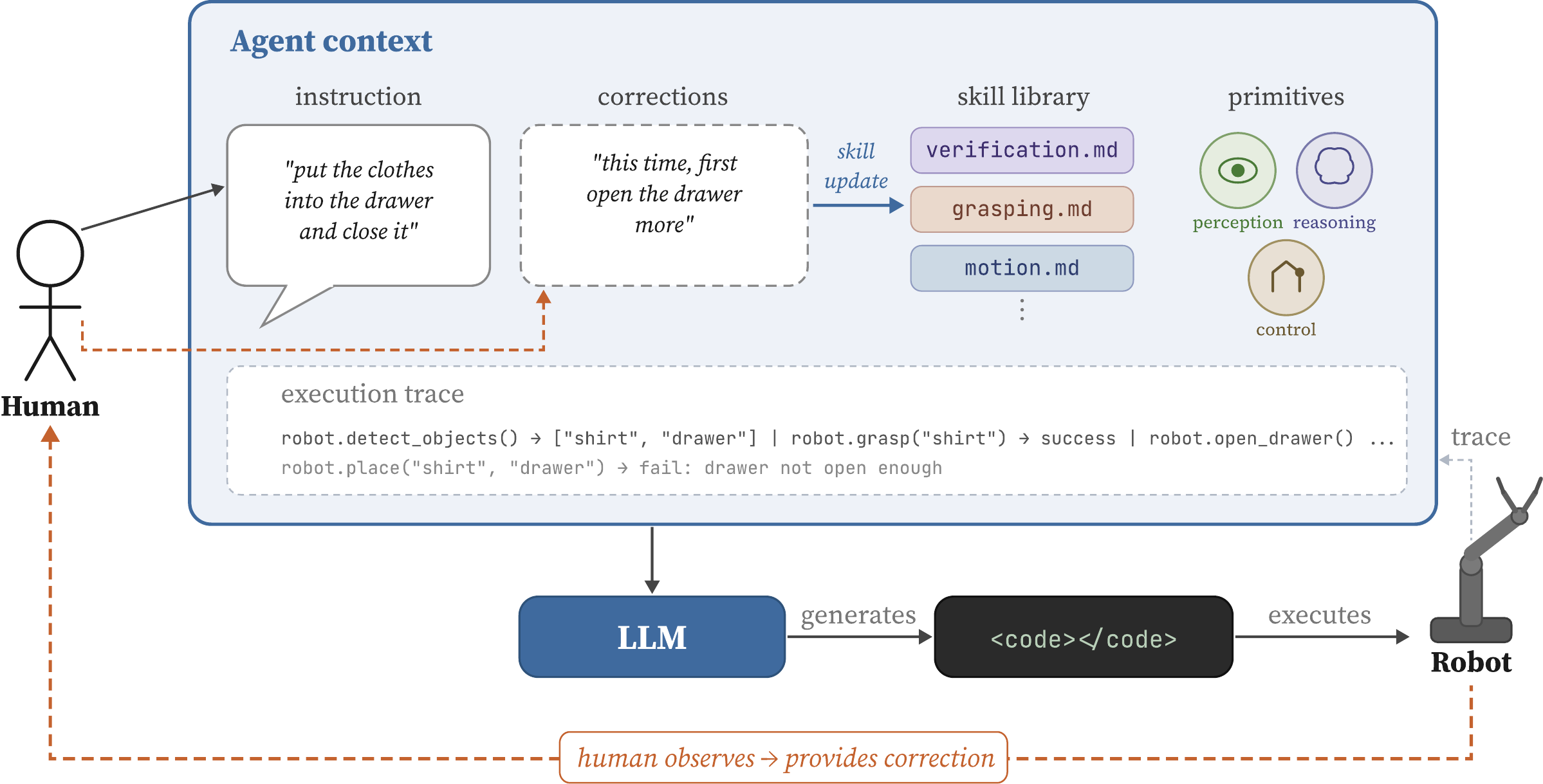}
        \caption{System overview of \method. The pipeline takes natural language input and synthesizes robot policies using a suite of tools including control primitives and perception modules. When execution fails or is suboptimal, a human observer provides a language correction, which both modifies the agent's current context for re-generation and \emph{updates the skill library} for use in future tasks. The agent also receives an execution trace from the robot, enabling self-diagnosis of failures (e.g., ``drawer not open enough'') to help localize code improvement from corrections.}
        \label{fig:system_overview}
    \end{figure}
	
    In contrast, modular approaches offer a path toward steerable autonomy by decoupling high-level reasoning from low-level execution, without requiring any expert demonstrations or robot-specific fine-tuning. Recent work has demonstrated that large-scale foundation models can act as agentic orchestrators, leveraging pre-trained vision and language priors to compose specialized tools for perception and control~\citep{liang2023codepolicieslanguagemodel, wang2025mosaicmodularfoundationmodels}. However, existing modular approaches are often limited by a \textit{specification gap}: the initial natural language instruction is frequently insufficient to capture the geometric and physical nuances of a task, leading to failures that the system cannot recover from autonomously. Recently, coding agents have demonstrated even non-experts can synthesize complex programs through iterative feedback in natural language. We envision capable robots that can adapt via the same principles: rather than training a black-box policy, users should be empowered to correct robots through natural language, as they would to a human or another agent.

    In this work, we propose \method (Agentic Robot Code with Human In ThE loop CorrecTions), a framework that treats policy acquisition as an interactive code synthesis task. \method does not require any robot-specific training data. Unlike prior work, \method treats the Large Language Model (LLM) not merely as a planner \citep{liang2023codepolicieslanguagemodel}, but as an orchestrator agent capable of synthesizing executable robot programs that invoke a suite of heterogeneous tools, including pretrained foundation models, open-vocabulary detectors, grasp generators, and motion planners. 

    The core of \method is a human-in-the-loop refinement process grounded in two key components: 1) \textbf{Policy Steering via Language Corrections:} Humans express intent and teach new concepts through language \cite{inbook, article, lupyan2025importantlanguagehumanlikeintelligence}. When a synthesized policy fails, the user provides a natural language correction (e.g., ``The grasp was too high, try picking it up from the rim''). Unlike direct code editing or manual reward-function tuning \citep{inproceedings, Sutton1998}, natural language is accessible to non-experts, mirroring how humans teach one another. Successful corrections are then distilled into a \textit{Skill Library}. This library serves as a form of long-term in-context learning, allowing the agent to ``remember'' prior corrections and reuse them as high-level primitives for future tasks, effectively building a repertoire of interpretable skills similar to the Voyager system \citep{wang2023voyageropenendedembodiedagent}. 2) \textbf{Agentic Refinement and Orchestration:} To ground these corrections in code updates, the agent analyzes \textit{program execution traces}. This allows the model to identify exactly which line of code or tool call led to the failure. \method can then re-build the program using the suite of tools. 

    We evaluate \method on a Franka Panda manipulator across $8$ diverse manipulation tasks, including common challenges like articulated and deformable objects. We demonstrate that \method significantly outperforms state-of-the-art VLA baselines $\pi_0$ and $\pi_{0.5}$, as well as robot program synthesis baseline Code As Policies \citep{liang2023codepolicieslanguagemodel}, particularly in tasks requiring multi-step reasoning. We also evaluate the value of humans in the loop by comparing \method performance with VLM-generated instead of HiTL corrections. Finally, we conduct an $N=6$ human evaluation where we demonstrate that the skill library enables generalization across task variations, showing \method enables rapid adaptation to new objects and scenes with just ``a few words.'' 

\section{Related Work}
\label{sec:citations}
\vspace{-1em}
    \paragraph{LLMs \& VLMs for Robot Code Synthesis} Many works have studied how to integrate their reasoning ability to generate embodied robot programs~\citep{liang2023codepolicieslanguagemodel, grannen2024vocalsandboxcontinuallearning, huang2022innermonologueembodiedreasoning, singh2022progpromptgeneratingsituatedrobot, wang2023demo2codesummarizingdemonstrationssynthesizing, liu2024okrobot}. Recent approaches have incorporated human feedback through visual demonstrations, uncertainty quantification, and language corrections~\citep{murray2025teaching, ren2023robotsaskhelpuncertainty, cui2023right}. However, these works typically aim to produce autonomous policies, and thus can not be improved upon execution. \citet{wang2025mosaicmodularfoundationmodels} leverage a modular architecture including foundation models to enable a user to interact with a robot through natural language commands. \citet{wu2023tidybot} use LLMs to both infer user preferences from summaries and do language-based planning. \citet{ahn2022saycan} use LLMs to encode high-level semantic knowledge about the world into a task planner that determines how to execute low-level skills. 
    \citet{fu2026capxframeworkbenchmarkingimproving} introduce a benchmark for improving coding agents for manipulation. 
\vspace{-1em}
    \paragraph{Robot Learning from Language Corrections} \citet{shi2024yellrobotimprovingonthefly} demonstrate that fine-grained language corrections can be used for improving high-level policies, but requires custom data collection and fine-tuning to produce the base policy. Other work in shared autonomy \citep{cui2023nototheright, grannen2024vocalsandboxcontinuallearning, wang2025mosaicmodularfoundationmodels} aims to produce a policy where the user can intervene with an utterance. More recently, work on policy steering \citep{chen2026steerablevisionlanguageactionpoliciesembodied} has shown that language can be used to improve pretrained models' performance for specific downstream tasks, but it does not incorporate corrections on the fly. 
	

\section{\method}
\label{sec:method}

    We present \method, an agentic framework for generating and adapting modular, interpretable \citep{ieee_transparency} robot manipulation policies from natural language.  Our approach, illustrated in \autoref{fig:system_overview}, accepts a language instruction as initial input and requires no robot demonstration data. It uses a combination of open source models, off-the-shelf APIs, and engineered components to process observations. Corrections allow a human supervisor to interactively compose complex skills. We first describe the problem of human-in-the-loop program synthesis, then provide a high-level overview of \method, and finally describe each component of the approach.

    \subsection{Overview \& Problem Formulation} 

    \method uses a pretrained LLM to produce robot policy code, mapping the user's input language instruction for the task to robot primitives that can control the robot's movement, respond to perceptual inputs, and query the robot's proprioceptive state. The LLM autoregressively generates functions until the synthesized code satisfies the instruction. The action space of the robot is determined by the control primitives, while the action space of the agent layer is governed by the tool calls that can be made in order to produce a policy. 

    \subsection{Language Guided Policy Synthesis}

    We use one input to initialize policy synthesis: a natural language instruction that describes the task. The language instruction is parsed and embedded into a system prompt which calls the LLM to generate code. The policy code is grounded in the scene by using a set of control and perception primitives as well as tools such as VQA, proprioception, or CLI commands to read the available skills. 
    Example system prompts and skills are included in \autoref{app:prompts}. After a detected failure or at the end of an episode, the human is queried for a correction. \autoref{fig:system_overview} provides an overview. 

    \subsection{Skill Library \& Memory}

    A challenge in human-in-the-loop robot learning is ensuring that corrections are not temporary. In many existing approaches, corrections are consumed once to update a policy and then effectively discarded, requiring the human to re-correct similar failures across sessions. The skill library addresses this by providing a structured mechanism for accumulating, retaining, and reusing knowledge. It has been applied in non-robotics domains~\citep{wang2023voyageropenendedembodiedagent}. 

    When a human provides a language correction during execution, \method synthesizes it into a skill: a concise, natural language rule or code pattern that encodes the adjustment, helping to bridge the gap between the human's intent and an executable robot policy. This skill is added to the skill library, a persistent collection that grows over the course of interaction. Before each subsequent code generation step, all skills in the library are loaded into the agent's context alongside the current task instruction and an execution trace from prior iterations. This means the model has access to the full history of learned corrections when synthesizing a new policy, allowing it to avoid previously identified failure modes without requiring a re-correction.

    \subsection{Tool Suite}
    \begin{wrapfigure}{r}{0.45\columnwidth}
        \centering
        \vspace{-12pt}
        \includegraphics[width=0.5\columnwidth]{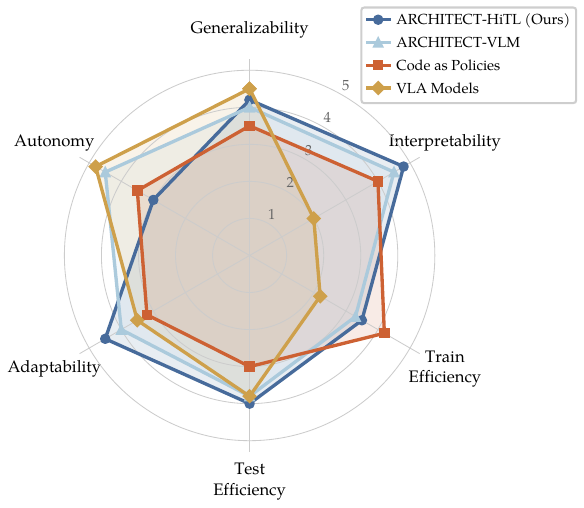}
        \caption{Comparison of \method, Code as Policies, and Vision-Language Action Models across $6$ axes of desirable traits for robot policies.}
        \label{fig:radar-comparison}
    \end{wrapfigure}

    The tools primarily consist of robot control, proprioception, and perception modules. They are defined to be simple and generic so as to not require task-specific assumptions, and can thus be used to compose flexible and diverse task logic. Functions are parameterized with a natural language input to ground the language corrections in feasible behavior. Together, these primitives form a complete set of tools that enable the robot with high-level reasoning (via parametrization in language) and low-level execution so that the synthesized policies can capture semantic intent and geometric precision. More details on the tool suite implementation are included in \autoref{app:tool_implementation}. 
\vspace{-1em}
	\paragraph{Control Tools} The control primitives are parametrized by geometric information from perception output. They represent the available action space of the robot. They are granular enough to define both global trajectory planning and fine-grained adjustments. They are adaptable in language space by way of \texttt{skills} which define common patterns and conventions in natural language, such as using a relative motion to adjust end effector position from the current pose. This enables the policy to mediate between the agent's natural language reasoning and the robot's geometric action space. 
\vspace{-1em}
    \paragraph{Perception Tools} The perception suite serves as a bridge between raw sensory data and actionable spatial information. \method processes observations from the environment using perception primitives that utilize off-the-shelf vision models, of which some are directly parameterized in natural language by way of text queries or descriptions. Additionally, we use grasp perception and object placement modules; both rely on input from an upstream language-parametrized module to be adapted by language corrections. These tools ground open-vocabulary concepts into the robot's physical frame and provide the necessary grounding to interact with the environment. 
\vspace{-1em}
    \paragraph{Proprioception Tools} The proprioception primitives allow the system to query the robot's current state in order to inform what action or logic should take place next for the policy to succeed.


\section{Experiments}

    In our experiments we seek to address 1) how well does \method compare to state of the art generalist policies and program synthesis methods and 2) to what extent do human corrections contribute to \method's performance versus VLM-generated feedback? From our human evaluation, we aim to answer 3) does \method enable high-level knowledge transfer to new scenes and objects and 4) do the skills that \method acquires from corrections in one task improve sample efficiency in new tasks by reducing the number of corrections required?

    \subsection{Setup} To evaluate \method we conduct a series of real-world experiments on a Franka Panda manipulator configured with the widely-adopted DROID setup~\citep{khazatsky2025droidlargescaleinthewildrobot}. The robot is equipped with a Robotiq 2F-85 gripper, a wrist-mounted ZED-mini stereo camera, and a ZED 2i for the scene camera. In our experiments, \method uses the Claude Opus 4.6 large language model~\citep{theC3} as the orchestrator.  We implement the control primitives using cuRobo for motion generation~\citep{sundaralingam2023curoboparallelizedcollisionfreeminimumjerk}, which allows our system to handle calculating a trajectory's reachability, inverse kinematics, and collision constraints. 

    \subsection{Task Suite} 
\begin{figure}[t]
        \centering
        \includegraphics[width=0.9\textwidth]{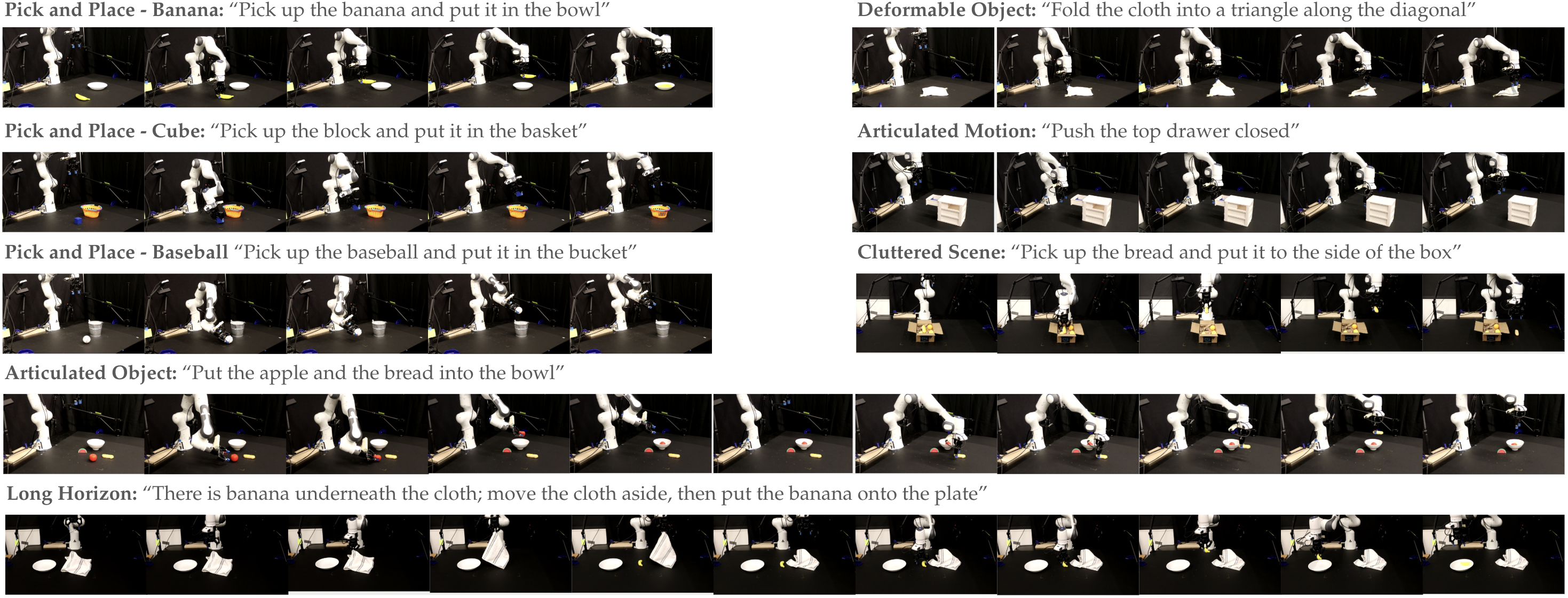}
        \caption{The 8 evaluation tasks and their challenge categories.}
        \label{fig:eval_tasks}
    \end{figure}

We evaluate our approach on a selection of 8 challenging tasks that variously require multi-step reasoning, non-prehensile manipulation, and interaction with deformable or articulated objects, shown in  \autoref{fig:eval_tasks}. The complete task instructions and success criteria are detailed in \autoref{app:eval_task_criteria}.

    \subsection{Benchmarking Evaluation Protocol} We evaluate our approach against the following baselines: Code As Policies~\citep{liang2023codepolicieslanguagemodel}, and the $\pi_{0}$ and $\pi_{0.5}$ VLA models from Physical Intelligence~\cite{black2026pi0visionlanguageactionflowmodel, intelligence2025pi05visionlanguageactionmodelopenworld}. Additionally, we conduct an ablation with \method using automatic VLM-generated corrections instead of human corrections, which we term \method-VLM. Since \method requires feedback to inform the skill library, we perform a maximum of 5 rollouts with an expert correction in each rollout, and evaluate the performance of our method using the final resulting policy. We evaluate each policy with $N=10$ episodes per task for a total of 400 rollouts for real-world benchmarking evaluation. We use the following metrics to benchmark performance, averaged across the 10 rollouts per task. Success rate ($SR$) is the most common standard metric and it is computed as a binary score determined by whether the task goals are met at the end of the episode. However, since $SR$ misses valuable signal on whether the policy partially succeeded, we also include two more metrics: goal condition recall ($GCR$), defined by \citet{murray2025teaching} as $GCR = \frac{(\# \text{ of goal conditions satisfied})}{(\text{total } \# \text{ of goal conditions required})}$ and normalized task progression $NTP$ defined by \citet{wang2026roboevalroboticmanipulationmeets} as the fraction of task stages that succeed according to an ordered task rubric, normalized by the number of steps. $GCR$ is well-suited for evaluating methods such as CaP which may execute independent unordered goals. Our rubric for measuring $NTP$ per task is included in \autoref{app:eval_task_criteria}. Object position/orientation and language instruction phrasing are perturbed for each new rollout to evaluate generalization of each method. 

    \subsection{Human Evaluation Protocol}

    In our human evaluation, we conduct a within-subjects experiment to isolate three effects: the value of human language corrections, skill library transfer to new scenes, and human effort amortization through the skill library. 
    Each participant completes two structurally similar manipulation tasks (Task 1: place a cup upright on a table; Task 2: place a box upright on a shelf) across six conditions that systematically toggle human corrections and skill library access. In each condition, the system generates a policy from the task instruction, executes it, and verifies each subgoal via VQA. In conditions with human-in-the-loop corrections, if subgoal verification results in failure, the participant observes the failure and provides a natural language correction (e.g., "grasp the cup from the top"), which the system uses to update a skill file and re-synthesize the policy. In the final round, skills accumulated during Task 1 are optionally carried forward to Task 2, enabling us to measure whether corrections on one task transfer to reduce failures and human effort on a related but distinct task.


    \begin{figure}[t]
        \centering
        \includegraphics[width=0.8\linewidth]{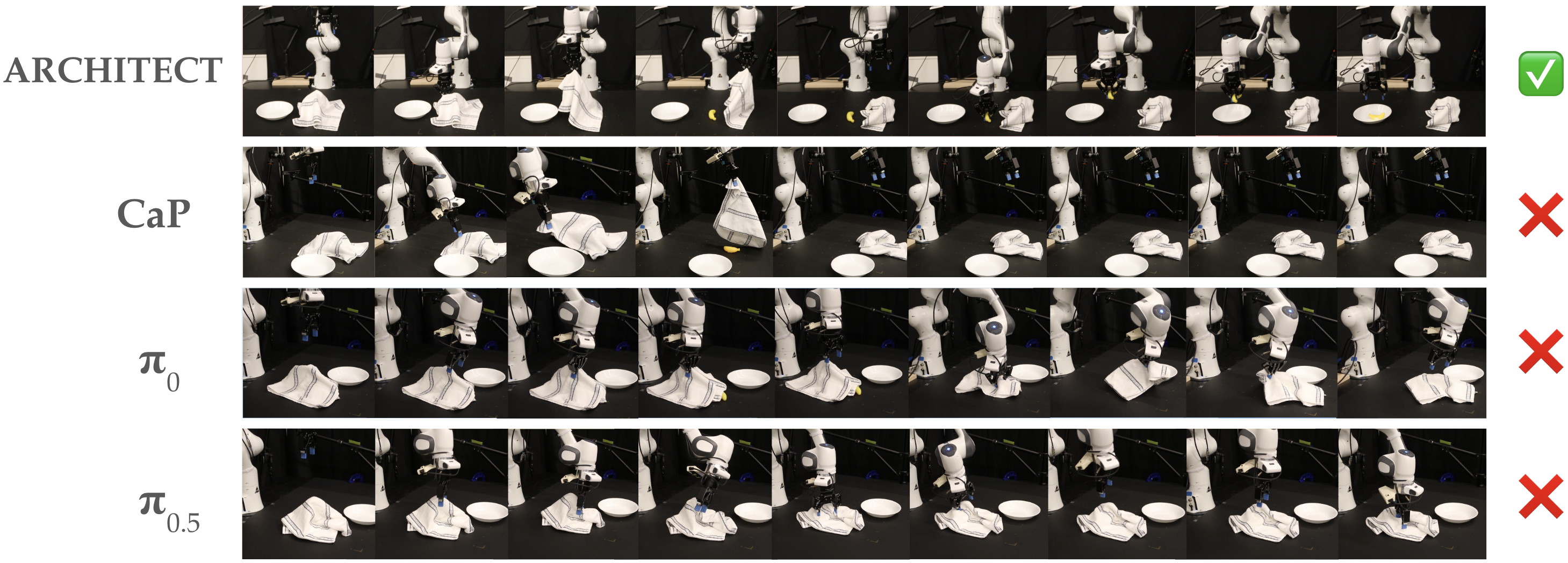}
        \caption{Comparison of \method performance against CaP, $\pi_0$, $\pi_{0.5}$ for a long horizon task (\textit{"There is banana underneath the cloth. Pick up the cloth and drop it aside, then pick and place the banana into the dish plate"}). \method completes the task, while others struggle: CaP and $\pi_0$ lift the cloth but cannot move it away from the banana. $\pi_{0.5}$ grasps the cloth but does not fully lift it.}
        \label{fig:comparison}
    \end{figure}

\section{Results}

    \subsection{\method outperforms generalist policies}

    Empirically, \method demonstrates strong capabilities over a range of manipulation tasks. We group the evaluation tasks into categories based on challenge type and report the results in \autoref{tab:results}. Per-task challenge conditions and evaluation rubric are detailed in \autoref{app:eval_task_criteria}. 
    Our method is able to utilize its available tools to solve challenging tasks. While most other methods were able to achieve pick-and-place tasks, \method demonstrated the ability to achieve more complex task categories with accuracy. 
    In comparison, VLAs $\pi_0$ and $\pi_{0.5}$ struggle to complete tasks with the same degree of complexity, evidenced by the low success rate for task categories outside of pick-and-place. 
    Additionally, our method achieved consistent performance within task families such as pick-place. In contrast, $\pi_0$ and $\pi_{0.5}$ exhibited high variance based on the target pick object, performing worse when presented with objects that are not well-represented in their training data~\citep{yang2026robolabhighfidelitysimulationbenchmark}. We also observed that $\pi_0$ and $\pi_{0.5}$ exhibited inconsistent performance within tasks based on phrasing of the language instruction, which was varied per rollout, whereas \method was robust to variations. 

    Furthermore, we find that \method achieves higher success than CaP, which similarly struggled to complete tasks other than pick-place. CaP demonstrates that LLMs can map language to robot programs, but assumes the initial generation is sufficient, thus is particularly brittle to perceptual ambiguity and task underspecification. 
    \method relaxes that assumption by treating policy synthesis as an iterative process with both autonomous self-verification (exec traces) and human-in-the-loop corrections. The performance gap reflects the difficulty of getting robot policies right on the first try from open-loop synthesis, even with strong LLMs and capable perception tools.
    In comparison, closing the loop with human corrections and execution feedback equips the agent with task-grounded context that a prompt alone cannot provide. 

    \begin{table}[t]
        \centering
        \caption{ Success Rate (SR), Goal Condition Recall (GCR), and Normalized Task Progression (NTP) across tasks. We compare \method against CaP~\citep{liang2023codepolicieslanguagemodel}, $\pi_0$~\citep{black2026pi0visionlanguageactionflowmodel}, $\pi_{0.5}$~\cite{intelligence2025pi05visionlanguageactionmodelopenworld}, and ablation \method-VLM.}
        \label{tab:results}
        \resizebox{\linewidth}{!}{%
        \setlength{\tabcolsep}{2.5pt}
        \scriptsize
        \begin{tabular}{ll lccccccccccccccc}
        \toprule
        & & & \multicolumn{3}{c}{CaP} & \multicolumn{3}{c}{$\pi_0$} & \multicolumn{3}{c}{$\pi_{0.5}$} & \multicolumn{3}{c}{\method-VLM} & \multicolumn{3}{c}{\method-HiTL (ours)} \\
        \cmidrule(lr){4-6} \cmidrule(lr){7-9} \cmidrule(lr){10-12} \cmidrule(lr){13-15} \cmidrule(lr){16-18}
        Task & Challenge & & SR & GCR & NTP & SR & GCR & NTP & SR & GCR & NTP & SR & GCR & NTP & SR & GCR & NTP \\
        \midrule
        Banana $\rightarrow$ plate & Pick-Place & & .20 & .52 & .52 & .80 & .90 & .90 & \textbf{.90} & .98 & .98 & .80 & .88 & .88 & .80 & .96 & .96 \\
        Block $\rightarrow$ basket & Pick-Place & & .40 & .52 & .52 & .40 & .44 & .44 & \textbf{1.0} & 1.0 & 1.0 & .70 & .80 & .80 & .70 & .82 & .82 \\
        Baseball $\rightarrow$ bucket & Pick-Place & & .60 & .66 & .66 & .00 & .16 & .16 & .20 & .50 & .50 & .40 & .62 & .62 & \textbf{1.0} & 1.0 & 1.0 \\
        Apple, Bread $\rightarrow$ bowl & Multi Pick-Place & & .10 & .59 & .47 & .40 & .64 & .64 & .40 & .65 & .65 & .00 & .49 & .39 & \textbf{.80} & .88 & .88 \\
        Pick bread from box & Clutter & & .10 & .38 & .38 & .40 & .40 & .40 & .40 & .44 & .44 & .00 & .24 & .24 & \textbf{.70} & .78 & .78 \\
        Close drawer & Articulated & & .00 & .00 & .00 & .10 & .23 & .23 & .00 & .17 & .17 & .40 & .60 & .60 & \textbf{.90} & .90 & .90 \\
        Fold cloth into triangle & Deformable & & .00 & .00 & .00 & .00 & .40 & .40 & .00 & .04 & .04 & .70 & .62 & .62 & \textbf{.80} & .95 & .95 \\
        Banana under cloth & Long Horizon & & .00 & .40 & .40 & .10 & .36 & .36 & .00 & .19 & .19 & .00 & .60 & .60 & \textbf{.80} & .80 & .60 \\
        \bottomrule
        \end{tabular}%
        }%
    \end{table}

    \subsection{\method amortizes human effort across tasks}

    A key practical claim of \method is that the skill library accumulated under HiTL corrections amortizes future human effort. We compare \method with an empty library against \method with the library populated from each participant's own previous task session. Mean queries per trial fell from $4.67$ to $0.83$ ($\Delta = 3.83$, $p = 0.036$), with $3/6$ participants requiring $0$ corrections.  
    This pattern distinguishes \method from the baselines. VLA policies would require additional demonstrations to specialize to a new task; they have no mechanism for incorporating a free-form correction during deployment, so the ``effort'' axis is paid up front in data collection time. CaP re-synthesizes code from scratch for each instruction and therefore does not amortize prior corrections across tasks. \method, in contrast, reuses skills authored under a \emph{different} task instruction at no additional cost, which is what drives the reduction in human queries, shown in \autoref{fig:sample_eff}. 

    \subsection{\method enables transfer to new scenes}

    Additionally, our human evaluation results isolate the value of the skill library \emph{without} any HiTL support at trial time. We find that \method enables transfer to novel scenes and objects for within-domain tasks. 
    Without skills, no participant succeeded on Task 2 ($\mathrm{SR} = 0/6$). With the Task 1 skill library loaded but no further correction loop, $\mathrm{SR}$ rose to $4/6$, a $\Delta$ visible in $4/6$ participants; the remaining $2/6$ tied at $0$. These results are shown in \autoref{fig:zeroshot_sr} and \autoref{fig:zeroshot_gcr}. The successes correlate with HiTL effort on the prior task. Of the three participants who gave only one Task 1 correction, two failed Task 2 zero-shot; all three who gave two or more corrections succeeded. This is consistent with a denser correction trace producing a more general skill library.
    Relative to baselines, both VLA policies and CaP treat a second task as a fresh problem: a VLA would need a new demonstration set, and CaP would re-prompt from the task instruction alone with no access to the strategies created during Task 1. \method's skill library makes prior human effort reusable across tasks without retraining or re-synthesis.

    \subsection{Human corrections address failure modes that VLM corrections cannot}
    
    Human corrections are able to address failure modes that are challenging to detect due to faulty or noisy sensors. For example, when depth estimation error causes the gripper to release an object too far above a surface, a human can immediately diagnose the gap between the intended and actual placement height, a discrepancy that is difficult to detect from camera images alone. Human corrections also encode physical intuitions about object properties that cannot be inferred from vision: a human might specify "close the gripper slowly so the baseball doesn't slip," a form of reasoning that is challenging for a VLM operating solely on image observations.
    
    \subsection{\method synthesizes policies from naive natural language corrections}

    Users were able to express corrections naturally and without robotics-specific jargon. Participant utterances included examples like "grasp the cup a little bit more to the left", or "keep moving down until the pasta box is touching the shelf" synthesizing policy updates that resulted in task success. A qualitative example of a synthesized skill is included in \autoref{app:skills}.

    \begin{figure}[t]
        \centering
        \begin{subfigure}[t]{0.32\linewidth}
            \includegraphics[width=\linewidth]{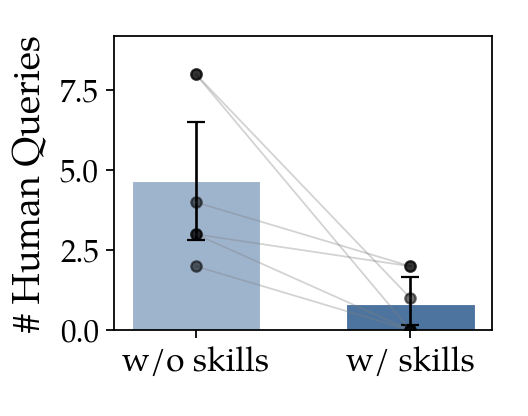}
            \caption{Sample efficiency.}
            \label{fig:sample_eff}
        \end{subfigure}\hfill
        \begin{subfigure}[t]{0.32\linewidth}
            \includegraphics[width=\linewidth]{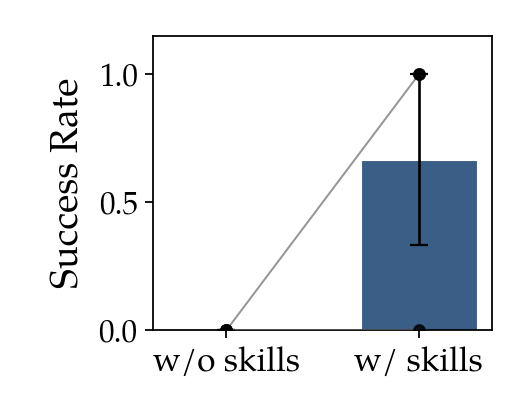}
            \caption{Transfer (success rate).}
            \label{fig:zeroshot_sr}
        \end{subfigure}\hfill
        \begin{subfigure}[t]{0.32\linewidth}
            \includegraphics[width=\linewidth]{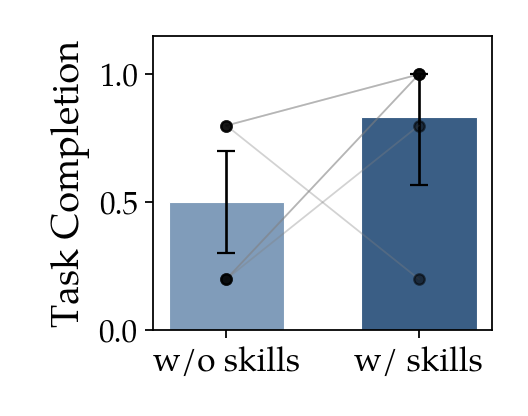}
            \caption{Transfer (task completion).}
            \label{fig:zeroshot_gcr}
        \end{subfigure}
        \vspace{3mm}
        \caption{\method results from human evaluation. \textbf{(a)} With a populated skill library, HiTL trials require fewer human queries at no cost to success rate. \textbf{(b, c)} Skills from Task 1 transfer to Task 2, raising both success rate and task completion.}
        \label{fig:t2_results}
    \end{figure}

\vspace{-2mm}
\section{Limitations and Future Work}
\vspace{-2mm}
\label{sec:limitations}

    \method is a capable system for interactively generating robot policies from code, but it has limitations which we plan to address in future work. Given that it is a modular approach, it is limited by the performance of the low-level primitives and underlying modules, such as the ability to determine grasps for objects, a known challenge in prior work~\citep{fang2023anygrasprobustefficientgrasp, murali2025graspgendiffusionbasedframework6dof, Yuan2023M2T2}. However, as new models are released and swapped in to our system, we believe its performance will improve. We address failure modes for \method compared to baselines in further detail in \autoref{app:failure_analysis}.
    \method is also limited by the number of corrections provided. This challenge was evident in the human evaluation, where noisiness in the grasp sampler meant that if a grasp succeeded and was thus not corrected, then \method would not necessarily have the skill in its library to correct future grasp failures. The two participants who provided only one correction on the first task did not complete the second task without additional corrections, indicating that the number of corrections matters for building a useful skill library. Future work could investigate performance with the number of corrections as well as study how correction quality affects the resulting policy. 

\vspace{-2mm}
\section{Conclusion}
\label{sec:conclusion}

    We present \method, an agentic framework for robot policy synthesis that closes the specification gap between natural language instructions and robust execution through human-in-the-loop code generation and a persistent, growing skill library. Across 8 real-world manipulation tasks spanning 6 challenge categories, \method outperforms state-of-the-art VLAs ($\pi_0$, $\pi_{0.5}$) and prior one-shot program synthesis method Code as Policies without any robot-specific training data, and its skill library amortizes human effort by enabling improved performance on novel in-domain tasks.

    Central to our approach is the role of the human-in-the-loop. While prior work demonstrates that scaling test-time computation through multi-model ensembles and autonomous multi-turn interaction can partially close the specification gap \citep{fu2026capxframeworkbenchmarkingimproving}, \method achieves comparable and often higher robustness through sparse human corrections. We find that a single natural language correction from a human supervisor often encodes the same diagnostic and physical information as multiple autonomous retry cycles. Furthermore, our ablation comparing human feedback to VLM-generated corrections reveals that human intuition remains critical for identifying subtle physical failure modes that current vision-language models fail to ground from pixels alone. 
    Importantly, \method demonstrates that human effort need not be ephemeral. Through the synthesis of a persistent skill library, the system effectively amortizes human supervision, enabling task transfer and a drastic reduction in human queries over time. As the field moves toward generalist robots, \method provides a steerable, data-efficient, and interpretable alternative to black-box learning, proving that in the path towards robot autonomy, a few words go a long way. 


\clearpage

\acknowledgments{The authors would like to thank members of the UW CSE robotics lab, including Marius Memmel and Jesse Zhang for the insights on setting up DROID and evaluating VLA policies. Thank you to Rulin Shao for the insights on coding agents and LLM research. Thank you to Mateo Guaman Castro and Gokul Swamy for the insights on experimental design. Lastly, thank you to Amazon Robotics for providing financial support for this project.}


\bibliography{example}  


\clearpage
\appendix

\section{Additional Experiments}

    In order to more comprehensively benchmark approaches, we compare \method against additional baselines including VLA models MolmoAct2 \citep{fang2026molmoact2actionreasoningmodels} and GR00T N1.7 \citep{nvidia2025gr00tn1openfoundation}, and LLM-based planner methods Inner Monologue \citep{huang2022innermonologueembodiedreasoning} and ProgPrompt \citep{singh2022progpromptgeneratingsituatedrobot}. The results are shown in Table \ref{tab:results_supp}. MolmoAct2 achieves the same performance as \method in one pick and place task (banana $\rightarrow$ plate) but does not outperform \method nor VLAs $\pi_0$, $\pi_{0.5}$ in any other task. Inner Monologue generally achieves higher performance than Code as Policies, but does not succeed over \method at any task. 
    ProgPrompt attains the highest success rate of any program synthesis method, but also does not achieve higher performance than \method. 
    GR00T N1.7 is the weakest baseline overall, registering a nonzero SR on banana $\rightarrow$ plate alone (0.40).

    Inner Monologue~\citep{huang2022innermonologueembodiedreasoning} is an LLM-based planner that closes the loop between planning and execution by feeding textual feedback (success detection, scene descriptions, and human responses) back into the LLM, which re-plans over a fixed library of pretrained skills. We select this baseline in order to compare another method that utilizes human input to disambiguate action selection. ProgPrompt~\citep{singh2022progpromptgeneratingsituatedrobot} prompts an LLM with a program specification of the environment (import statements over available actions, object lists, and example programs) to generate situated task plans as executable code with state checks.  
    
    MolmoAct2~\citep{fang2026molmoact2actionreasoningmodels} is an open action reasoning model that performs intermediate depth-aware spatial reasoning before acting, coupling an embodied-reasoning VLM backbone with a flow-matching continuous action expert. We choose this model in our benchmark because it allows comparison against another recent VLA architecture in addition to the $\pi$ models. GR00T N1.7~\citep{nvidia2025gr00tn1openfoundation} is an open cross-embodiment VLA combining a vision-language backbone with a diffusion action head, pretrained on a mixture of real robot trajectories, simulation, and large-scale egocentric human video. 
    MolmoAct2 exhibited strong performance on the banana pick-and-place task, but struggled with the remaining pick-and-place tasks, suggesting fragility to object type. Additionally, the model struggled with more complex tasks, often misunderstanding the prompt entirely, e.g. pressing the top of the cabinet rather than approaching the drawer to close it. A recurring challenge was premature disengagement from extended prompts: in the multi-object pick and place, it would successfully move one of the objects, but would withdraw before attempting a second, and in the cluttered scene, it would pick the wrong object and fail to return for the correct one. 

\begin{table}[]
    \centering
    \caption{ Success Rate (SR), Goal Condition Recall (GCR), and Normalized Task Progression (NTP) across tasks. We compare \method against Inner Monologue (IM)~\citep{huang2022innermonologueembodiedreasoning}, ProgPrompt~\citep{singh2022progpromptgeneratingsituatedrobot}, MolmoAct2~\citep{fang2026molmoact2actionreasoningmodels}, and GR00T N1.7~\citep{nvidia2025gr00tn1openfoundation}}
    \label{tab:results_supp}
    \resizebox{\linewidth}{!}{%
    \setlength{\tabcolsep}{2.5pt}
    \scriptsize
    \begin{tabular}{ll lccccccccccccccc}
    \toprule
    & & & \multicolumn{3}{c}{Inner Monologue} & \multicolumn{3}{c}{ProgPrompt}
    & \multicolumn{3}{c}{MolmoAct2} & \multicolumn{3}{c}{GR00T N1.7} & \multicolumn{3}{c}{\method (ours)} \\
    \cmidrule(lr){4-6} \cmidrule(lr){7-9} \cmidrule(lr){10-12} \cmidrule(lr){13-15} \cmidrule(lr){16-18}
    Task & Challenge & & SR & GCR & NTP & SR & GCR & NTP 
    & SR & GCR & NTP & SR & GCR & NTP & SR & GCR & NTP \\
    \midrule
    Banana $\rightarrow$ plate & Pick-Place & & .50 & .90 & .90 & \textbf{.80} & .92 & .92 
    & \textbf{.80} & .92 & .92 & .40 & .64 & .64 & \textbf{.80} & .96 & .96 \\
    Block $\rightarrow$ basket & Pick-Place & & .50 & .76 & .76 & .60 & .68 & .68 
    & .30 & .64 & .64 & .00 & .20 & .20 & \textbf{.70} & .82 & .82 \\
    Baseball $\rightarrow$ bucket & Pick-Place & & .50 & .84 & .84 & .60 & .64 & .64 
    & .00 & .34 & .34 & .00 & .10 & .10 & \textbf{1.0} & 1.0 & 1.0 \\
    Apple, Bread $\rightarrow$ bowl & Multi Pick-Place & & .30 & .51 & .49 & .10 & .43 & .38 
    & .10 & .43 & .43 & .00 & .05 & .05 & \textbf{.80} & .88 & .88 \\
    Pick bread from box & Clutter & & .00 & .00 & .00 & .10 & .34 & .34 
    & .20 & .32 & .32 & .00 & .06 & .06 & \textbf{.70} & .78 & .78 \\
    Close drawer & Articulated & & .00 & .00 & .00 & .00 & .00 & .00 
    & .00 & .00 & .00 & .00 & .00 & .00 & \textbf{.90} & .90 & .90 \\
    Fold cloth into triangle & Deformable & & .00 & .00 & .00 & .00 & .00 & .00 
    & .00 & .08 & .08 & .00 & .00 & .00 & \textbf{.80} & .95 & .95 \\
    Banana under cloth & Long Horizon & & .00 & .30 & .30 & .00 & .15 & .14 
    & .00 & .47 & .47 & .00 & .24 & .22 & \textbf{.80} & .80 & .60 \\
    \bottomrule
    \end{tabular}%
    }%
\end{table}

\section{Language Ablation}

    We conduct an ablation study to determine the effect of perturbing the language instruction on method performance. For this ablation, we pick the best-performing VLA from our evaluation, $\pi_{0.5}$, and vary the language instruction based on paraphrasing, length, distractors, and specification. The exact language instructions are included below. In contrast to our benchmarking evaluation, the target object position and orientation are held constant for the language ablation in order to isolate the effect of the language instruction alone. The results are displayed in Table \ref{tab:lang_ablation}. Overall, \method demonstrates robustness against variations in the language instruction while $\pi_{0.5}$ is sensitive to changes, particularly in task underspecification, overspecification, and paraphrasing. 

    Notably, $\pi_{0.5}$ showed significant sensitivity to prompt specificity. When given underspecified prompts, $\pi_{0.5}$ exhibited undirected movement before eventually orienting toward the task, most clearly seen in the "Put the apple away" task, where the arm moved at random without targeting the apple. A similar degradation was observed with an overly specified prompt, suggesting that $\pi_{0.5}$ struggles when clear task intent is either absent or obscured. 

    \begin{table}[h]
        \centering
        \captionof{table}{Language instruction perturbation ablation on the \textit{Apple $\rightarrow$ pan} task. We compare $\pi_{0.5}$~\citep{intelligence2025pi05visionlanguageactionmodelopenworld}, the best-performing VLA from our main evaluation, against \method under perturbations in the instruction paraphrasing, length, introducing distractors, and level of specification. Target object position and orientation are held constant across all conditions.}
        \label{tab:lang_ablation}
        \resizebox{\linewidth}{!}{%
        \setlength{\tabcolsep}{3pt}
        \scriptsize
        \begin{tabular}{lp{7.2cm} cccccc}
        \toprule
        & & \multicolumn{3}{c}{$\pi_{0.5}$} & \multicolumn{3}{c}{\method (ours)} \\
        \cmidrule(lr){3-5} \cmidrule(lr){6-8}
        Perturbation & Instruction & SR & GCR & NTP & SR & GCR & NTP \\
        \midrule
        Canonical & Put the red apple in the pan. & 1.0 & 1.0 & 1.0 & 1.0 & 1.0 & 1.0 \\
        \midrule
        \multirow{4}{*}{Paraphrase (synonym)}
        & Grab the apple and set it on the pan. & 1.0 & 1.0 & 1.0 & 1.0 & 1.0 & 1.0 \\
            & Take the apple and place it onto the pan. & .00 & .20 & .20 & 1.0 & 1.0 & 1.0 \\
        & Move the apple onto the pan. & .00 & .20 & .00 & 1.0 & 1.0 & 1.0 \\
        & Lift the apple and rest it on the pan. & 1.0 & 1.0 & 1.0 & 1.0 & 1.0 & 1.0 \\
        \midrule
        Paraphrase (passive) & The apple should be picked up and placed on the pan. & 1.0 & 1.0 & 1.0 & 1.0 & 1.0 & 1.0 \\
        \midrule
        Paraphrase (order) & Onto the pan, place the apple. & .00 & .20 & .20 & 1.0 & 1.0 & 1.0 \\
        \midrule
        Paraphrase (ambiguity) & Relocate the fruit so that it sits on the pan. & .00 & .20 & .20 & 1.0 & 1.0 & 1.0 \\
        \midrule
        Length (short) & apple on pan & 1.0 & 1.0 & 1.0 & 1.0 & 1.0 & 1.0 \\
        \midrule
        Length (medium) & Pick up the apple from the table and place it onto the pan sitting nearby. & .00 & .20 & .20 & 1.0 & 1.0 & 1.0 \\
        \midrule
        Length (long) & There is a apple resting on the surface in front of you. Pick that apple up, then move it over and place it down onto the pan that is also on the surface, so the apple ends up on the pan. & 1.0 & 1.0 & 1.0 & .00 & .80 & .80 \\
        \midrule
        Distractor (typos) & pickup teh appel an put it on the frying pan & .00 & .20 & .20 & 1.0 & 1.0 & 1.0 \\
        \midrule
        \multirow{2}{*}{Underspecification}
        & Put the apple away. & .00 & .20 & .20 & 1.0 & 1.0 & 1.0 \\
        & Move the apple. & .00 & .20 & .20 & .00 & .60 & .60 \\
        \midrule
        Overspecification & Pick up the red apple and put it on the white pan; approach from directly overhead, keep the gripper aligned with the apple's long axis, and release only once it is centered over the pan. & .00 & .20 & .20 & 1.0 & 1.0 & 1.0 \\
        \bottomrule
        \end{tabular}%
        }%
    \end{table}

\section{Complete Evaluation Task List \& Criteria}
\label{app:eval_task_criteria}

    We use the following rubric per task to evaluate task progression. One point is awarded per step.

    \textbf{Pick and place (banana-plate):} the robot picks up a banana and places it onto a plate.
    \begin{itemize}
        \item End effector positioned above banana 
        \item Gripper securely grasps the banana
        \item Robot lifts the grasped banana clear of the workspace surface
        \item Gripper with grasped banana is positioned above plate 
        \item Gripper releases the banana, banana is entirely within plate
    \end{itemize}

    \textbf{Pick and place (block-basket):} the robot picks up a block and places it into a basket.
    \begin{itemize}
        \item End effector positioned above block 
        \item Gripper securely grasps the block
        \item Robot lifts the grasped block clear of the workspace surface
        \item Gripper with grasped block is positioned above basket 
        \item Gripper releases the block, block is entirely within basket
    \end{itemize}

    \textbf{Pick and place (baseball-bucket):} the robot picks up a baseball and places it into a bucket. Compared to other pick-and-place tasks, this evaluates performance on a spherical, low-friction object.
    \begin{itemize}
        \item End effector positioned above baseball 
        \item Gripper securely grasps the baseball
        \item Robot lifts the grasped baseball clear of the workspace surface
        \item Gripper with grasped baseball is positioned above bucket 
        \item Gripper releases the baseball, baseball is entirely within bucket
    \end{itemize}

    \textbf{Multi-step pick and place (red apple and bread into bowl):} the robot sequentially picks two distinct objects, a red apple followed by a slice of bread, from the workspace which also contains a watermelon slice. It then places them into the target bowl. This task evaluates the policy's ability to chain two complete pick-and-place subroutines into a single instruction without intermediate human intervention.
    \begin{itemize}
        \item End effector positioned above the red apple
        \item Gripper securely grasps the red apple
        \item Robot lifts the grasped apple clear of the workspace surface
        \item Gripper with grasped apple is positioned above the bowl
        \item Gripper releases the apple, apple is entirely within the bowl
        \item End effector positioned above the bread
        \item Gripper securely grasps the bread
        \item Robot lifts the grasped bread clear of the workspace surface
        \item Gripper with grasped bread is positioned above the bowl
        \item Gripper releases the bread, bread is entirely within the bowl
    \end{itemize}

    \textbf{Pick in clutter (bread):} the robot retrieves a slice of bread from inside a cluttered box (containing other items such as corn, orange, lemon, and packing paper) and places it outside the box.
    \begin{itemize}
        \item End effector positioned above the bread
        \item Gripper securely grasps the bread (without co-grasping the wrapper)
        \item Robot lifts the grasped bread clear of the box rim
        \item Gripper with grasped bread is positioned outside the footprint of the box
        \item Gripper releases the bread onto the workspace surface beside the box
    \end{itemize}

    \textbf{Fold cloth:} the robot grasps the bottom corner of a square cloth and folds it in half diagonally to produce a triangular shape by aligning the bottom corner with the diagonally opposite top corner.
    \begin{itemize}
        \item End effector aligned with the bottom corner of the cloth
        \item Gripper securely grasps the cloth corner
        \item Robot lifts the grasped corner clear of the workspace surface
        \item Grasped corner is positioned above the geometric center of the cloth
        \item Grasped corner is positioned at the pre-drop pose aligned with the opposite (top) corner
        \item Gripper releases the cloth, resulting in a triangular fold with the two corners within tolerance
    \end{itemize}

    \textbf{Close drawer:} the robot closes an open cabinet drawer by applying a pushing motion to the drawer front. 
    \begin{itemize}
        \item End effector aligned with the drawer front face
        \item Robot approaches the drawer by applying a forward push
        \item Drawer is fully closed (flush with cabinet body)
    \end{itemize}

    \textbf{Long horizon (banana under cloth):} the robot first removes a cloth that occludes a banana, then performs a pick-and-place of the now-visible banana into a dish plate. This task evaluates long-horizon behavior with an obligatory occlusion-resolution prerequisite before the target manipulation can begin.
    \begin{itemize}
        \item End effector aligned with the cloth for a pre-pick pose
        \item Gripper securely grasps the cloth
        \item Robot lifts the grasped cloth clear of the workspace surface
        \item Robot moves the cloth aside, away from the banana
        \item Gripper releases the cloth such that the banana is fully visible and accessible for grasping
        \item End effector positioned above the banana 
        \item Gripper securely grasps the banana
        \item Robot lifts the grasped banana clear of the workspace surface
        \item Gripper with grasped banana is positioned above the dish plate 
        \item Gripper releases the banana, banana is entirely within the dish plate
    \end{itemize}

\section{Human Evaluation Task Details}
\label{app:human_eval}

    We chose the two tasks for the human evaluation based on the similar challenges that could lead to failures requiring corrections. For Task 1, "place the cup upright on the table", the robot must properly grasp the cup so that it does not slip, lift it high enough so that it does not hit the table, and place it gently so that it does not topple over. Similarly, for Task 2, "place the pasta box upright on the shelf", the robot must properly execute the grasp, ensure that the box does not move the shelf while lifting it, and gently lower it into the placement position so that it does not fall off the shelf. Photos of each task are shown below. 

    \begin{figure}[h]
        \centering
        \begin{subfigure}[t]{0.32\linewidth}
            \includegraphics[width=\linewidth]{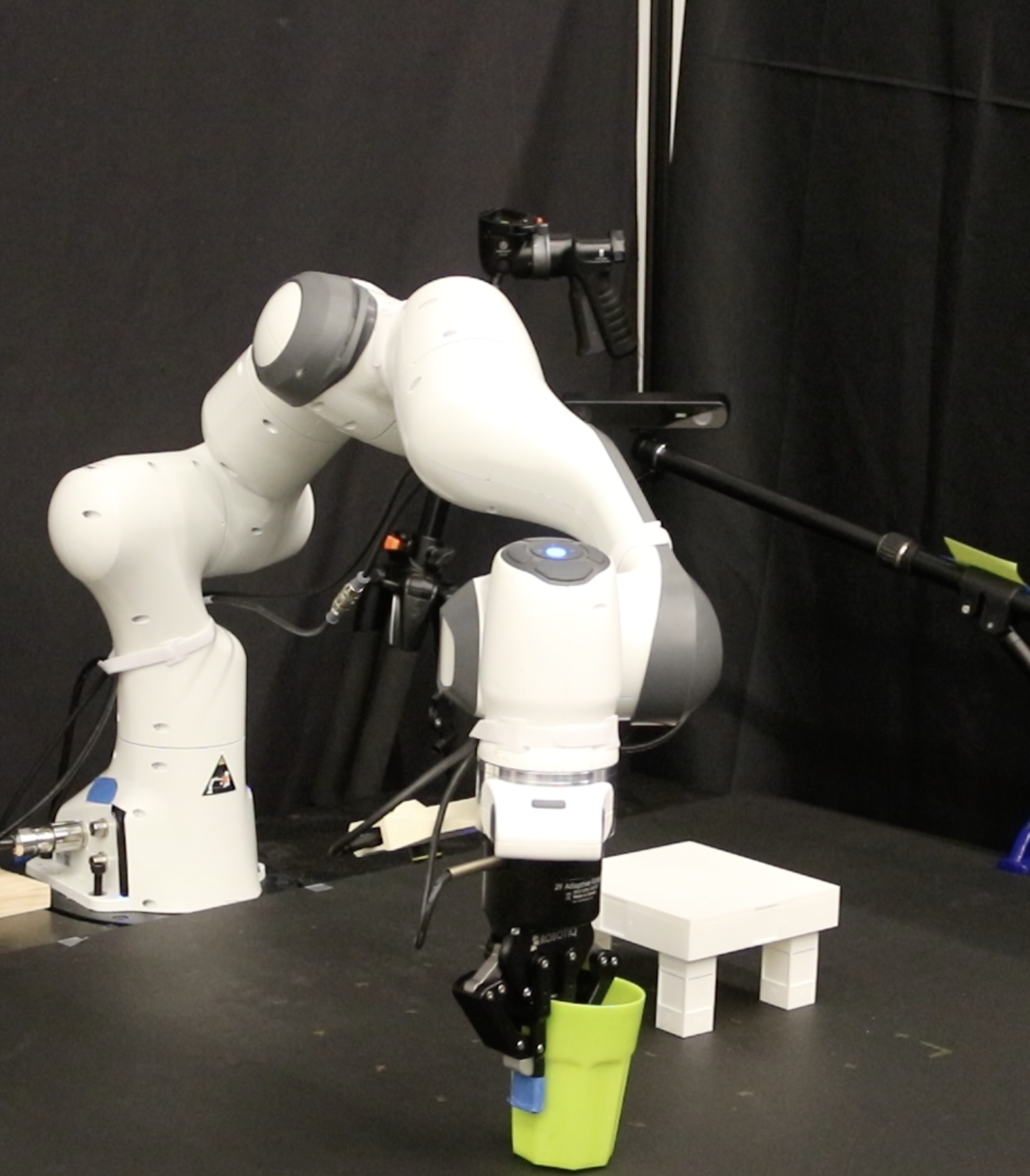}
            \caption{Task 1 start.}
            \label{fig:sample_eff}
        \end{subfigure}\hfill
        \begin{subfigure}[t]{0.307\linewidth}
            \includegraphics[width=\linewidth]{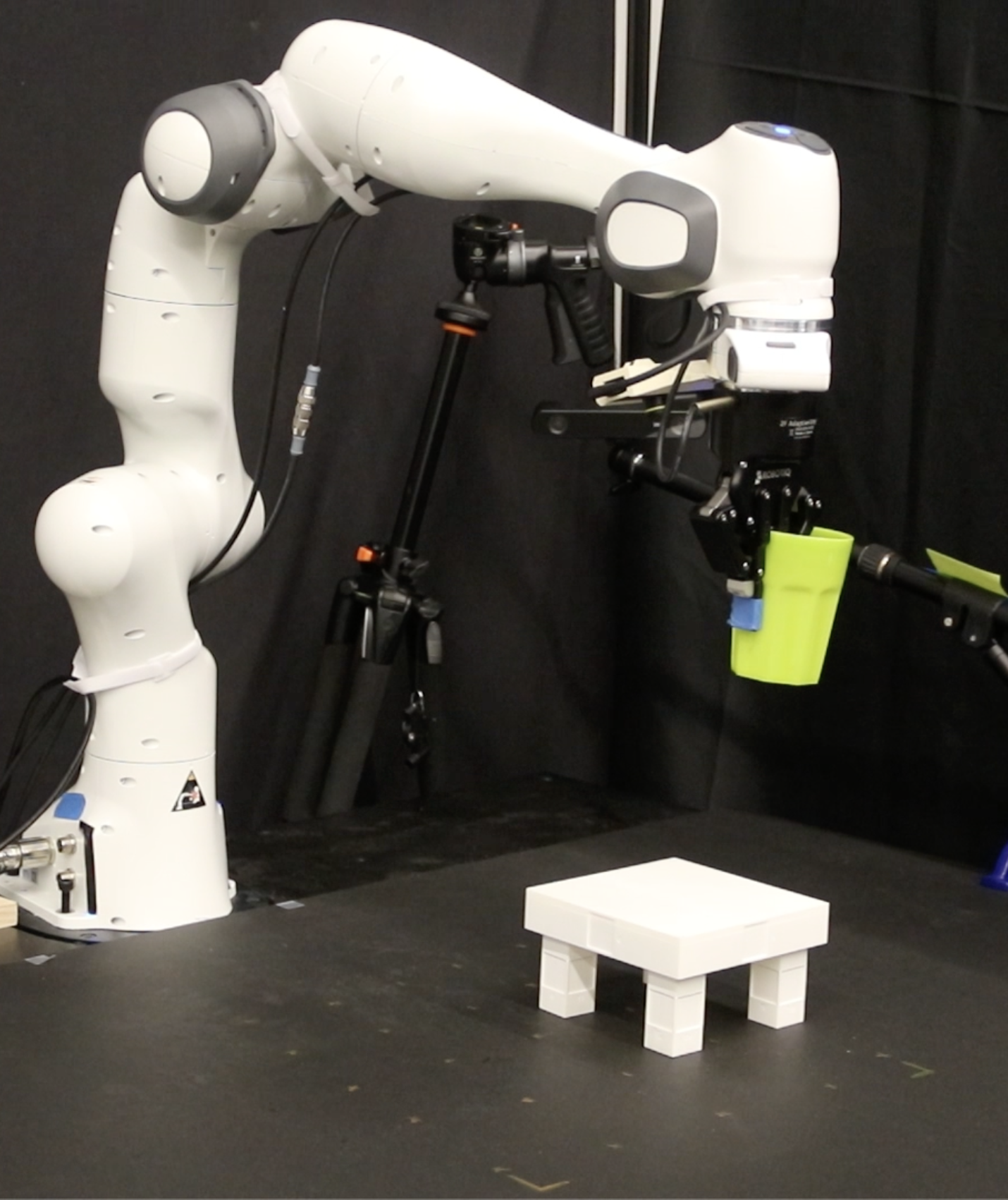}
            \caption{Task 1 middle.}
            \label{fig:zeroshot_sr}
        \end{subfigure}\hfill
        \begin{subfigure}[t]{0.315\linewidth}
            \includegraphics[width=\linewidth]{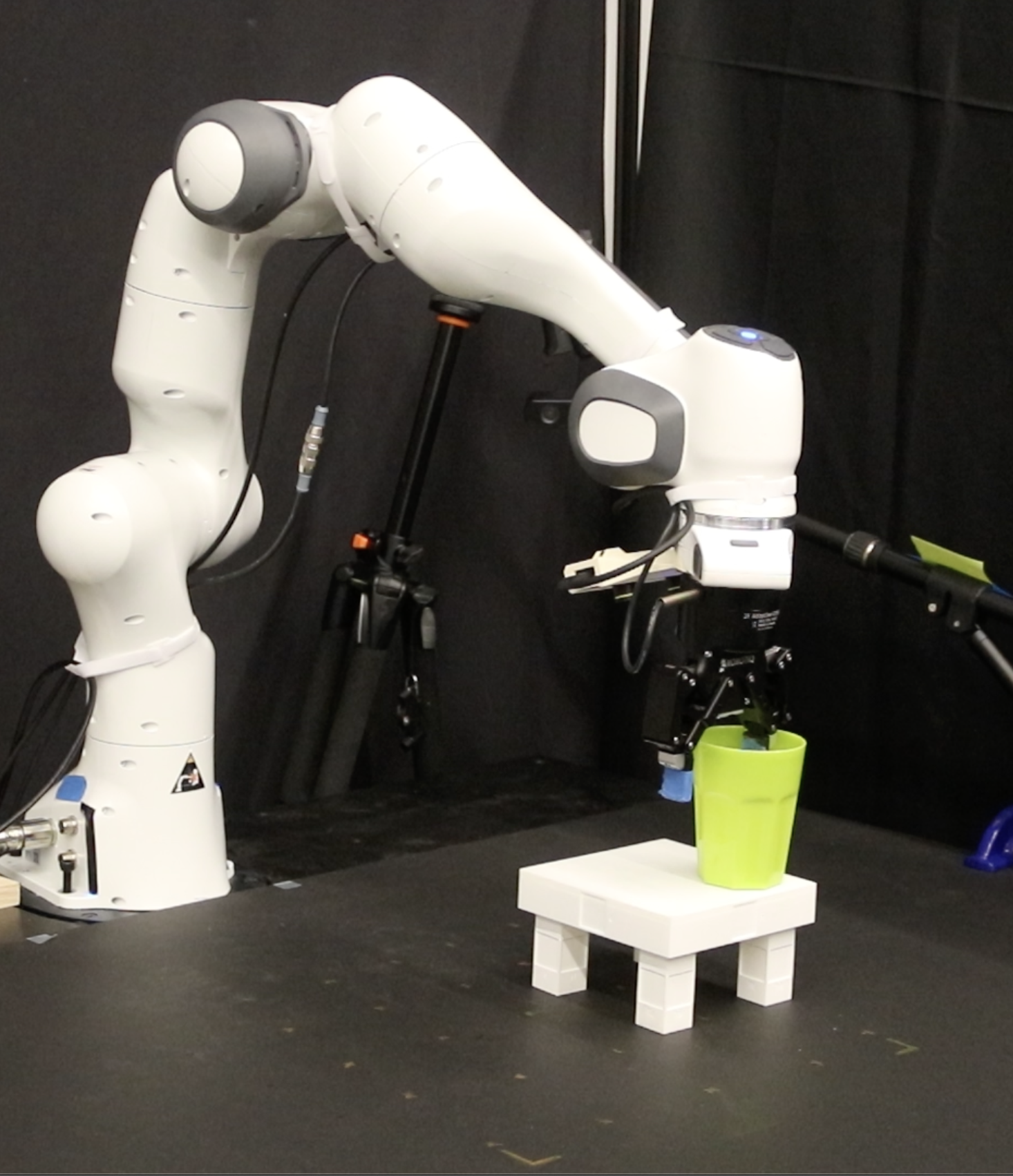}
            \caption{Task 1 end.}
            \label{fig:zeroshot_gcr}
        \end{subfigure}
        \vspace{3mm}
        \caption{Task 1 from human evaluation. The instruction is "place the cup upright on the table".}
        \label{fig:he_task1}
    \end{figure}

    \begin{figure}[h]
        \centering
        \begin{subfigure}[t]{0.30\linewidth}
            \includegraphics[width=\linewidth]{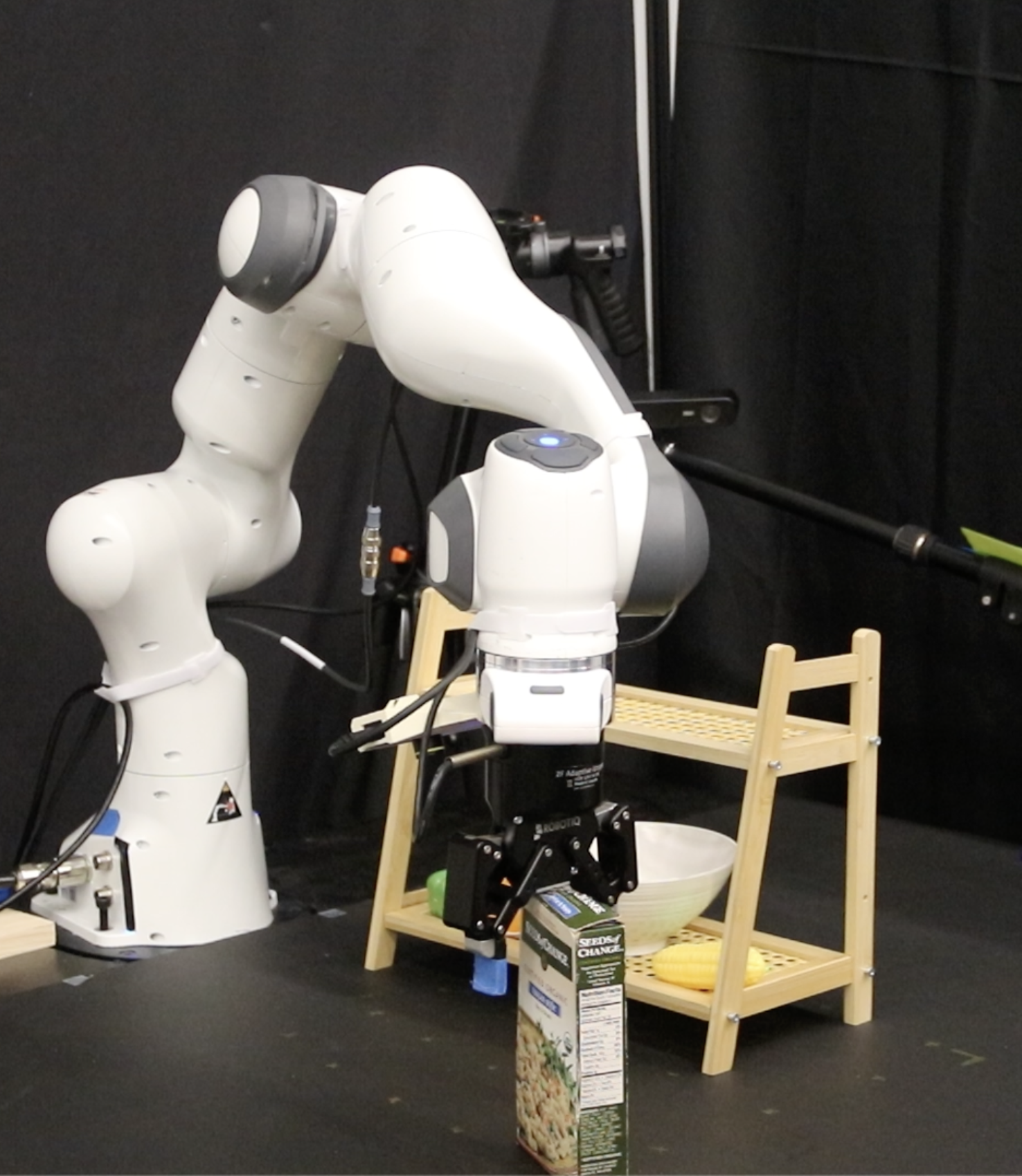}
            \caption{Task 2 start.}
            \label{fig:sample_eff}
        \end{subfigure}\hfill
        \begin{subfigure}[t]{0.31\linewidth}
            \includegraphics[width=\linewidth]{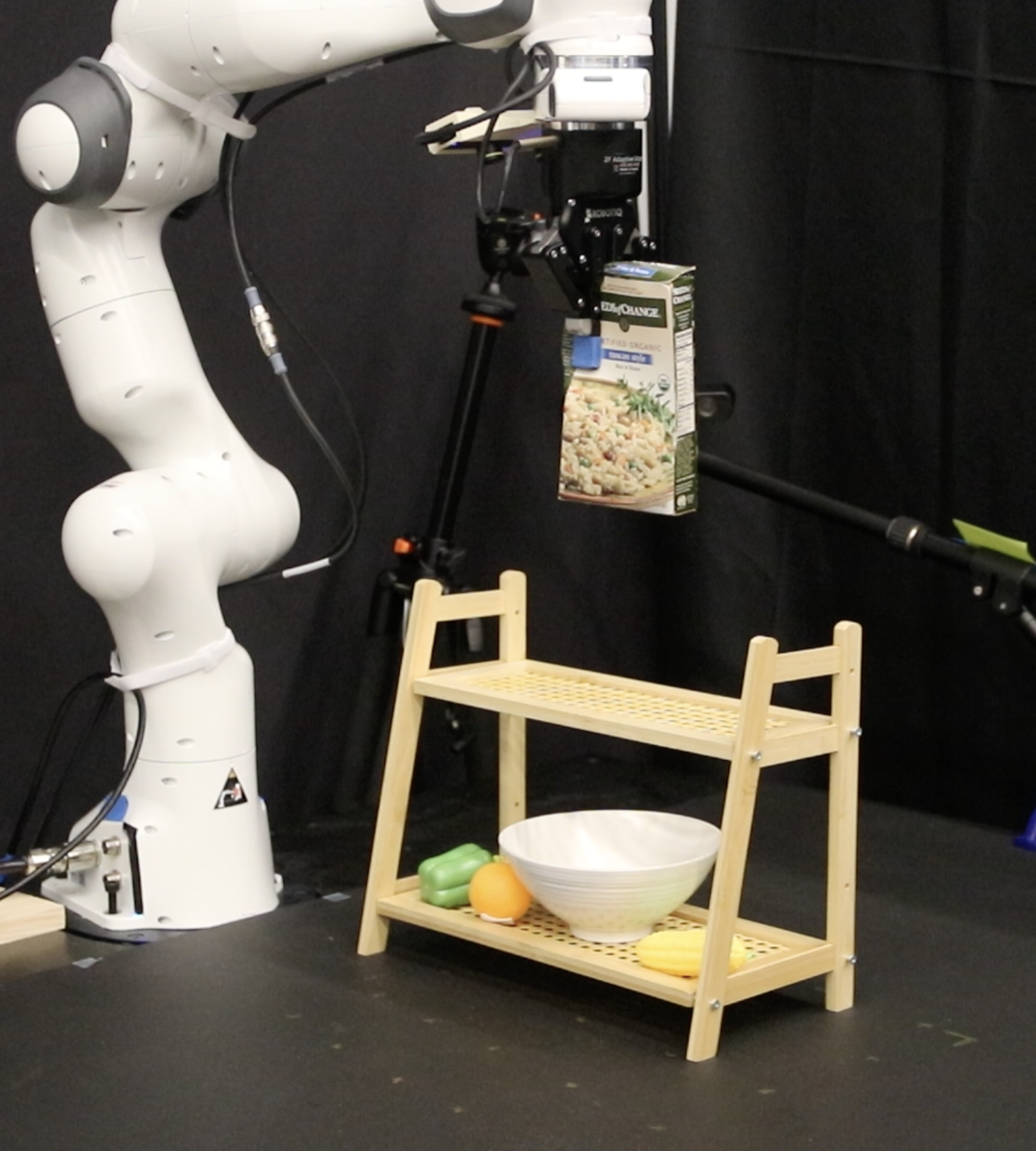}
            \caption{Task 2 middle.}
            \label{fig:zeroshot_sr}
        \end{subfigure}\hfill
        \begin{subfigure}[t]{0.315\linewidth}
            \includegraphics[width=\linewidth]{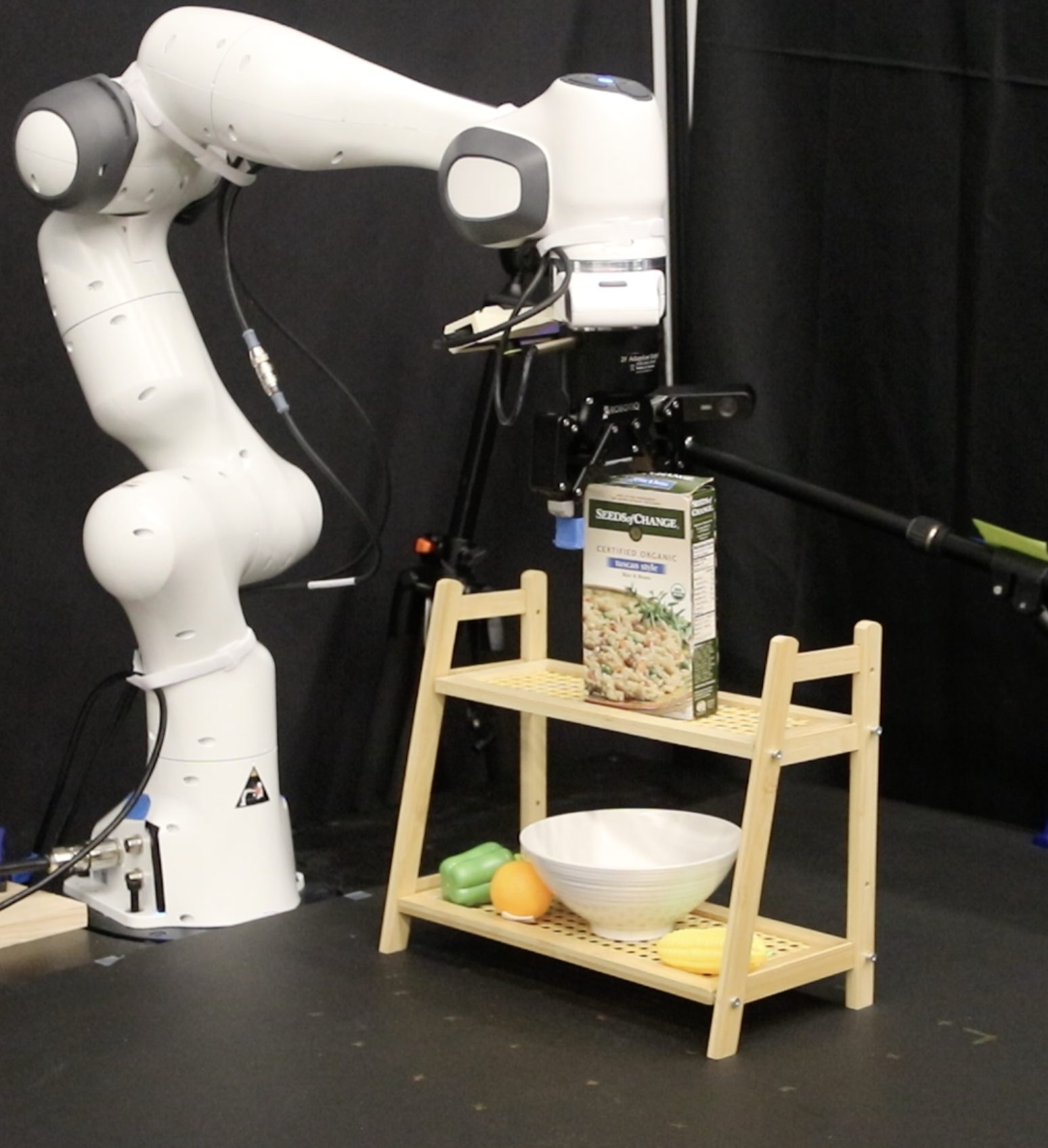}
            \caption{Task 2 end.}
            \label{fig:zeroshot_gcr}
        \end{subfigure}
        \vspace{3mm}
        \caption{Task 2 from human evaluation. The instruction is "place the pasta box upright on the shelf".}
        \label{fig:he_task2}
    \end{figure}

\section{Failure Analysis}
\label{app:failure_analysis}

    The primary failure mode of policies produced by \method is grasping. This is a common challenge in robotic manipulation policies and is due to the off-the-shelf grasp generation module we chose. Additionally, our method struggles to identify accurate placement poses when depth estimation is inaccurate, e.g. for dropping items into a bucket, it may raise the end effector too high over the bucket surface. This is also a limitation of the underlying module but can be addressed by user corrections. Lastly, \method sometimes faces motion planning errors when cuRobo is unable to produce a collision-free trajectory for the given target pose. 

    In contrast, the primary failures of VLA policies included sensitivity to prompt formulation, object-type dependence, and poor spatial reasoning. Models often failed to account for vertical clearance constraints: for example, ignoring the height of barriers in pick-and-place tasks, or insufficiently lifting deformable objects before setting them aside. Performance also varied substantially with object type, as reflected in the deviation between pick and place success rates across tasks. Finally, when required to distinguish between objects or follow a specific procedure to accomplish the task, models frequently defaulted to an alternative approach that precluded completion of the task, e.g. when asked to close a drawer by pushing it, a model might instead grasp the lip of the drawer, preventing full closure.

\section{Example Prompts}
\label{app:prompts}

    \begin{promptbox}[title={System Prompt}]
        \begin{lstlisting}[gobble=8]
        You are an agentic program synthesizer for a Franka Panda + Robotiq 2F-85. 
        Your goal is to decompose the task into named primitives and produce a clean, readable top-level program that calls them. Programs that are flat lists of low-level API calls are NOT acceptable - every meaningful sub-operation must be its own primitive.

        MANDATORY workflow - follow every step in order:

        1. Call list_primitives() to see the full robot API and any primitives already \
        registered this session.
        
        2. Call query_robot_state() and get_scene_description() to ground your plan in
        the current robot state and scene. Then call the appropriate perception \
        functions up-front before writing the program:
          - detect_objects() - ONLY for objects that must be GRASPED (picked up). \
        Do NOT call detect_objects() for placement targets (surfaces, containers, etc.).
          - get_placement_pose(base_text_prompt, target_text_prompt) - for placement \
        targets. The base is the surface/container you are placing onto, and the \
        target is the object being placed.
        For example, "pick up the banana and put it in the frying pan" requires \
        detect_objects("banana") for the grasp and \
        get_placement_pose("frying pan", "banana") for the placement - NOT \
        detect_objects("frying pan").
        
        3. Identify sub-operations. For EACH sub-operation that involves more than one \
        API call (e.g. "grasp object", "place on shelf", "navigate to position"), you MUST \
        call write_primitive(name, code, docstring) to register it before using it in the \
        program. Do not inline multi-step logic into the top-level program.
        
        4. Only after all primitives are registered, call submit_program(code) with a \
        short top-level program that reads as a sequence of named primitive calls.
        
        Tool reference:
        - list_primitives        - see all available robot API + session-registered primitives
        - read_primitive(name)   - get source of a session primitive
        - query_robot_state      - read current EE pose, joints, gripper width
        - get_scene_description  - get VQA scene description
        - write_primitive(name, code, docstring) - register a named sub-operation
        - run_ros_command(command) - ROS CLI introspection
        - submit_program(code)   - deliver the final top-level program
        
        Rules for write_primitive():
        - Code must use only functions from list_primitives() output or previously registered primitives
        - No imports; no references to undefined names
        - One primitive = one named sub-task; keep it focused
        
        Rules for submit_program():
        - Top-level program should be a short, readable sequence of primitive calls with brief comments
        - No imports; no inline multi-step logic
        - You MUST call submit_program() - do not output the program as text
        
        Verification by default - for any sub-operation whose success is not 
        observable from proprioception alone (grasps, placements, opening doors 
        drawers / lids), prefer the VQA-verified-retry pattern: perform the action, 
        ask `get_vqa_response` a yes/no question keyed on the intended outcome, and 
        on a negative answer adjust the relevant parameter and retry once. 
        Open-loop sequences are fine for trivial steps (`reset_robot`, home, 
        free-space moves). See `verification.md` in the domain knowledge below for 
        the concrete code template and failure-handling rules.
        
        Refer to the domain knowledge sections below for guidance on grasping, VQA, 
        verification-by-retry, motion constraints, and ROS introspection.
        \end{lstlisting}
    \end{promptbox}

\section{Skill Library Examples}
\label{app:skills}

    The following skill is synthesized by the coding agent with only the following correction text as input: \textit{"when moving the pasta box down onto the shelf, keep moving down until the pasta box is touching the shelf. Do not ungrasp if the pasta box is so high that it would fall, and do not end the task without having ungrasped the pasta box."}

    The skill contains usage guidelines, an example of the correction grounded in code, and rules specified as function parameters for the tool calls. 
    
    \begin{promptbox}[title={Skill: Guarded Place on Surface}]

        {\large\bfseries Guarded Place on Surface}
        
        \medskip
        {\bfseries\normalsize When to use}
        
        \smallskip
        When placing a grasped object onto a surface (shelf, table, tray, etc.)
        where the exact height of the surface relative to the end-effector is
        uncertain. Instead of moving to a fixed pose and releasing---which risks
        dropping the object from too high---use a guarded descent to ensure the
        object is in contact with the surface before opening the gripper.
        
        \medskip
        {\bfseries\normalsize How}
        
        \smallskip
        After moving the end-effector to the approximate placement pose (which
        may be slightly above the actual surface), perform a \textbf{guarded move
        downward} along the z-axis with a generous distance limit and a force
        threshold. This ensures the object descends until it physically contacts
        the surface. Only then should you open the gripper.
        
        \begin{lstlisting}
        # Step 1: Move to the approximate placement pose above the surface
        place_pose = {
            "position": place_position,
            "orientation": top_down_orientation
        }
        move_ee_to_pose(place_pose)
        
        # Step 2: Guarded descent
        move_ee_guarded(axis="z", distance=-0.25, force_threshold=5.0)
        
        # Step 3: Release the object
        set_gripper_width(0.085)
        
        # Step 4: Retract upward
        move_ee_to_rel_pose({"x": -0.05, "y": 0, "z": 0.1})
        \end{lstlisting}
        
        \medskip
        {\bfseries\normalsize What to watch for}
        
        \begin{itemize}[leftmargin=1.5em, itemsep=3pt]
          \item \textbf{Never ungrasp before contact.} If the placement pose
            leaves the object hovering above the surface, releasing will cause
            the object to fall. Always confirm surface contact via guarded move
            before opening the gripper.
          \item \textbf{Never end the task while still grasping.} The task is
            not complete until the gripper has been opened. Always include the
            \texttt{set\_gripper\_width()} call after the guarded descent.
          \item \textbf{Use a generous \texttt{distance} parameter} in
            \texttt{move\_ee\_guarded}. The distance is a maximum---the robot
            stops early when force is detected. A value like \texttt{-0.25} is
            safe for most placements.
          \item \textbf{Force threshold tuning:} A threshold of ${\sim}5.0$\,N
            works well for rigid surfaces. Increase slightly for soft surfaces.
          \item \textbf{Verify placement afterward} with VQA to confirm the
            object is correctly positioned and oriented.
        \end{itemize}
        
    \end{promptbox}

\section{Additional Details on Tool Suite Implementation}
\label{app:tool_implementation}

    \method exposes a suite of primitives across three categories -- control, perception, and proprioception -- summarized in Tables~\ref{tab:primitives-control}, \ref{tab:primitives-perception}, and \ref{tab:primitives-proprioception}.
    
    \bigskip
    \noindent
    \begin{minipage}{\linewidth}
    \centering
    \captionof{table}{\method tool suite primitives: Control}
    \label{tab:primitives-control}
    \begin{tabular}{@{}p{3.3cm} p{3.6cm} p{5.9cm}@{}}
        \toprule
        \textbf{Name} & \textbf{Input} & \textbf{Description} \\
        \midrule
        \texttt{set\_gripper\_width()} & \texttt{width} (float, 0.0--0.085\,m); optional: \texttt{speed}, \texttt{force} (0--255) & Commands the Robotiq 2F-85 gripper to a target aperture via linear position mapping; polls feedback until the target is reached or an object is detected \\
        \addlinespace
        \texttt{move\_ee\_to\_pose()} & \texttt{target\_pose} (position + quaternion); optional: \texttt{guard}, \texttt{force\_threshold}, \texttt{guard\_axes} & Moves end effector to an absolute pose via cuRobo trajectory planning over WebSocket; optionally enables per-axis contact guarding with automatic reflex recovery \\
        \addlinespace
        \texttt{move\_ee\_to\_rel\_pose()} & \texttt{delta\_position} (\{x, y, z\} in metres, EE frame); optional: \texttt{guard}, \texttt{force\_threshold}, \texttt{guard\_axes} & Applies a relative displacement in the end-effector frame; delta is rotated to the base frame and executed as a planned trajectory \\
        \addlinespace
        \texttt{move\_ee\_guarded()} & \texttt{axis} (x/y/z), \texttt{distance} (m), \texttt{force\_threshold} (N, optional) & Single-axis guarded move in the EE frame; cancels the trajectory and freezes the equilibrium pose on contact \\
        \addlinespace
        \texttt{execute\_waypoints()} & \texttt{waypoints} (list of \{position, quaternion, gripper\_action\}) & Sequentially executes a list of end-effector waypoints, republishing each as an equilibrium pose until convergence; optionally triggers a gripper open/close action at each waypoint \\
        \addlinespace
        \texttt{rotate\_wrist()} & \texttt{angle\_deg} (float), \texttt{direction} (cw/ccw) & Rotates the wrist about the tool Z-axis by composing an axis-angle rotation with the current EE orientation; delegates execution to \texttt{move\_ee\_to\_pose} \\
        \addlinespace
        \texttt{reset\_robot()} & None & Returns the robot to a predefined home configuration via trajectory planning and opens the gripper to full aperture \\
        \bottomrule
    \end{tabular}
    \end{minipage}
    
    \bigskip
    \begin{longtable}{@{}p{3.3cm} p{3.6cm} p{5.9cm}@{}}
    \caption{\method tool suite primitives: Perception}
    \label{tab:primitives-perception} \\
    \toprule
    \textbf{Name} & \textbf{Input} & \textbf{Description} \\
    \midrule
    \endfirsthead
    \multicolumn{3}{c}{\tablename\ \thetable\ -- \textit{Continued from previous page}} \\
    \toprule
    \textbf{Name} & \textbf{Input} & \textbf{Description} \\
    \midrule
    \endhead
    \midrule
    \multicolumn{3}{r}{\textit{Continued on next page}} \\
    \endfoot
    \bottomrule
    \endlastfoot
        \addlinespace
        \texttt{detect\_objects()} & \texttt{text\_prompt}, \texttt{camera} (wrist/scene); optional \texttt{x/y/z\_offset} & Chains Grounded SAM2 \citep{ren2024groundedsamassemblingopenworld} segmentation with AnyGrasp \citep{fang2023anygrasprobustefficientgrasp} grasp sampling; returns the highest-scored grasp pose transformed to the robot base frame via TF \\
        \addlinespace
        \texttt{get\_placement\_pose()} & \texttt{base\_text\_prompt}, \texttt{target\_text\_prompt}; optional \texttt{x/y/z\_offset} & Predicts a placement pose for a target object onto a support surface via AnyPlace \citep{zhao2025anyplacelearninggeneralizedobject}; segments both objects, builds point clouds, and returns the placement pose in the robot base frame \\
        \addlinespace
        \texttt{get\_vqa\_response()} & \texttt{prompt} (natural language), \texttt{camera} (wrist/scene); optional \texttt{max\_tokens}, \texttt{temperature} & Visual question answering via GPT-5.4 \citep{singh2026openaigpt5card}; captures a frame from the selected camera and returns a natural-language answer to the query \\
        \addlinespace
        \texttt{get\_keypoints()} & \texttt{text\_prompt}, \texttt{task} (natural language), \texttt{camera} (wrist/scene) & Extracts task-relevant keypoints via DIFT \citep{tang2023emergentcorrespondenceimagediffusion} diffusion features; captures RGB-D, looks up the camera-to-base transform, and returns keypoint coordinates in the robot base frame \\
        \addlinespace
        \texttt{get\_keypoints\_\allowbreak trajectory()} & \texttt{text\_prompt}, \texttt{task} (natural language), \texttt{camera} (wrist/scene) & Computes staged waypoint trajectories via a ReKep-based planner \citep{huang2024rekepspatiotemporalreasoningrelational} over DIFT keypoints; returns gripper-action-annotated waypoints grouped by manipulation stage \\
        \addlinespace
        \texttt{verify\_grasp()} & \texttt{object\_name} (string) & Two-stage grasp verification: first applies a proprioceptive gate on gripper width to reject obvious failures (fully open, fully closed, or below a minimum threshold), then queries a VLM with the wrist-camera image to confirm whether the named object is held \\
        \addlinespace
        \texttt{get\_object\_mask()} & \texttt{text\_prompt}, \texttt{camera} (wrist/scene) & Open-vocabulary segmentation via Grounded SAM2 \citep{ren2024groundedsamassemblingopenworld}; captures RGB-D from the selected camera and returns a lossless NPZ payload containing the RGB image, depth map, binary mask, and camera intrinsics \\
        \addlinespace
        \texttt{get\_grasp\_config()} & \texttt{npz\_base64} (segmentation mask), \texttt{camera} (wrist/scene); optional \texttt{x/y/z\_offset} & Standalone grasp sampling via AnyGrasp \citep{fang2023anygrasprobustefficientgrasp} from a pre-computed segmentation mask; transforms the highest-scored grasp pose to the robot base frame \\
    \end{longtable}
    
    \bigskip
    \noindent
    \begin{minipage}{\linewidth}
    \centering
    \captionof{table}{\method tool suite primitives: Proprioception}
    \label{tab:primitives-proprioception}
    \begin{tabular}{@{}p{3.3cm} p{3.6cm} p{5.9cm}@{}}
        \toprule
        \textbf{Name} & \textbf{Input} & \textbf{Description} \\
        \midrule
        \texttt{get\_current\_ee\_pose()} & None & Returns the current end-effector pose (position + quaternion) by extracting and converting the \texttt{O\_T\_EE} homogeneous transform from \texttt{FrankaState} \\
        \addlinespace
        \texttt{get\_current\_joints()} & None & Returns position, velocity, and torque for all seven joints from \texttt{FrankaState} \\
        \addlinespace
        \texttt{get\_gripper\_width()} & None & Returns the current gripper aperture width and per-finger positions in metres from Robotiq 2F-85 feedback \\
        \bottomrule
    \end{tabular}
    \end{minipage}

\end{document}